%% file: main.tex
\documentclass[11pt,dvipsnames]{article}
\usepackage[margin=1.1in]{geometry}
\usepackage{macros}
\usepackage{amsmath}
\usepackage{amssymb}
\usepackage{amsthm}
\usepackage{mathtools}
\usepackage{graphicx}
\usepackage{hyperref}
\usepackage[capitalize]{cleveref}
\usepackage{bm}
\usepackage{xcolor}
\usepackage{algorithm}
\usepackage[noend]{algpseudocode}
\usepackage{booktabs}
\usepackage{pgfplots}
\usepackage{multirow}
\usepackage[labelfont=bf]{subcaption}
\usepackage{microtype}

\usepackage[normalem]{ulem}

\newcommand{\mask}{\mathbf{m}}

\newcommand{\unrolledap}{\textbf{\texttt{UnrolledAP}}}
\newcommand{\trainedpg}{\textbf{\texttt{TrainedPG}}}
\newcommand{\prdeep}{\textbf{\texttt{prDeep}}}
\newcommand{\denoiser}{\textbf{\texttt{DRUNet}}}
\newcommand{\sol}{x_{\star}}

\theoremstyle{remark}

\usepackage{authblk}

\title{\textbf{Deeply unrolled alternating projections for phase retrieval}}
\author[1]{Jiayi Yao}
\author[1,5]{Vasileios Charisopoulos}
\author[2,3,4,5,6]{Rebecca Willett}

\affil[1]{Electrical \& Computer Engineering, University of Washington}
\affil[2]{Department of Computer Science, University of Chicago}
\affil[3]{Committee on Computational and Applied Mathematics, University of Chicago}
\affil[4]{Data Science Institute, University of Chicago}
\affil[5]{NSF-Simons National Institute for Theory and Mathematics in Biology}
\affil[6]{NSF-Simons AI-Institute for the Sky (SkAI)}

\date{}

\pgfplotsset{compat=newest}
\pgfplotsset{/pgfplots/error bars/y dir={both}}
\pgfplotsset{/pgfplots/error bars/y explicit}
\pgfplotsset{/pgfplots/error bars/error bar style={solid, thick}}
\tikzset{
    mredstyle/.style = {
        very thick,
        densely dashed,
        color=blue!80!black,
        mark=*,
        mark options={solid},
    },
    prdeep/.style = {
        very thick,
        densely dashed,
        color=red!80!black,
        mark=square*,
        mark options={solid},
    },
    proxgrad/.style = {
        very thick,
        densely dashed,
        color=green!40!black,
        mark=triangle*,
        mark options={solid},
    }
}

\allowdisplaybreaks

\begin{document}

\maketitle

\begin{abstract}
	Deep unrolling methods for inverse problems augment traditional reconstruction algorithms with learned priors to improve their performance.
	These methods have been developed under the implicit assumption that the geometry of the measurement operator does not change between training and deployment; when this is no longer the case, the quality of obtained reconstructions can degrade significantly.
	In this paper, we argue that this is not an inherent limitation of the deep unrolling paradigm and can be mitigated by a careful choice of the underlying iterative algorithm.\
	We propose a deep unrolling framework for the phase retrieval problem that augments the alternating projections method with a learned denoising step.
	Our method obtains high-quality reconstructions without the need for retraining in settings where conventional model-based alternatives struggle, such as when the dynamic range of the measurement operator changes significantly at test time.
\end{abstract}

\section{Introduction}
\label{sec:introduction}
\input{sections/introduction}

\section{Method}
\label{sec:method}

\input{sections/method}

\section{Experiments}
\label{sec:experiments}
\input{sections/experiments}

\input{sections/conclusion}

\section*{Acknowledgements}
RW acknowledges support from NSF DMS-2023109, AFOSR DOD FA9550-24-1-0327, and the NSF-Simons AI Institute for the Sky (SkAI) (via grants NSF AST-2421845 and Simons Foundation MPS-AI-00010513).

\bibliographystyle{IEEEbib}
\bibliography{main}

\end{document}

%% file: sections/introduction.tex
The phase retrieval problem---namely, reconstructing a signal from phaseless measurements---has diverse applications in X-ray crystallography, optics and remote sensing.
Formally, the unknown signal $\sol \in \mathbb{C}^{d}$ is observed indirectly via $m$ magnitude-only measurements $y \in \mathbb{R}^m$ satisfying:
\begin{equation}
    y_{i} \approx \abs{\ip{a_{i}, x_{\star}}},
    \label{eq:phase-retrieval-measurements}
\end{equation}
where $a_{i} \in \mathbb{C}^d$ are known \emph{design vectors} and $\abs{\cdot}$ indicates the complex magnitude operator.
Common sampling schemes include \emph{coded diffraction patterns},
masked Fourier measurements resulting in an isometric measurement operator $\mathcal{A}$:
\begin{equation}
y = \abs{\mathcal{A}(x)}, \;\; \text{for} \;\; \mathcal{A}(x) = \begin{bmatrix}
    \mathcal{F}(\mask_{1} \circ x) \\
    \vdots \\
    \mathcal{F}(\mask_{k} \circ x)
\end{bmatrix},
\label{eq:forward-model}
\end{equation}
where $\mathcal{F}$ denotes the (unitary) DFT,
$\circ$ is the elementwise (Hadamard) product, and $\mask_i \in \mathbb{C}^d$ are masks satisfying
\begin{equation}
    \mask_{i} = [\exp(i \theta_j)]_{j=1}^d, \;\;
    \text{where} \;\;
    \theta_{j} \iid \mathrm{Unif}([0, 2\pi)).
    \label{eq:mask-generation}
\end{equation}

\subsection{Data-driven recovery methods} \label{sec:subsec:recovery-methods}
Phase retrieval problems are typically ill-posed and computationally challenging to solve (see~\cite{FS20} for a survey); therefore, modern methods harness additional structural information to obtain high-quality reconstructions, with data-driven reconstruction methods~\cite{OJM+20} (including so-called \textit{deep image priors}~\cite{UVL18,MME19}, \emph{plug-and-play}~\cite{VBW13,KMW17,KBBW23} (PnP), and \emph{deep unrolling}~\cite{MLE21} methods) becoming the dominant paradigm.
Deep unrolling, which is the focus of this paper, replaces the regularization step of iterative reconstruction methods (\emph{e.g.,} proximal gradient methods) with a learnable regularizer --- commonly implemented as a neural network --- that
is tuned end-to-end by minimizing an aggregate reconstruction metric, after a fixed number of iterations, over a training dataset of $(x_\star,y)$ pairs.

\subsection{Our contribution} \label{sec:subsec:our-contribution}
The success of the deep unrolling paradigm hinges on the implicit assumption that the measurement operator is fixed across train and test time.
Nevertheless, practice dictates that imaging optics can be recalibrated, inducing a form of distribution shift (e.g., by introducing new sensors).
In such cases, end-to-end learned methods can apply miscalibrated updates that overfit to the geometry of the measurement operator at train time~\cite{gilton2021model}.
Even simple instantiations of the paradigm, including learnable step sizes~\cite{AMMG19},
can suffer under this setting. 

In this paper, we argue that this brittleness is \emph{not intrinsic} to deep unrolling but rather a consequence of the choice of \emph{which} iterative algorithm to unroll. In particular, we show that the classical alternating projections method~\cite{GS72,Fienup82}, embedded in the deep unrolling framework with a RED-type regularizer~\cite{REM17}, automatically adapts to the conditioning of the underlying measurement operator $\mathcal{A}$ at test time in regimes where other algorithms such as proximal gradient degrade sharply.
Our proposal is motivated by the equivalence of alternating projections with preconditioned (sub)gradient descent, described in detail in~\cref{sec:subsec:alternating-projections-with-red}.

\smallskip

\noindent \textbf{Distribution Shift.}
To model the shift between measurement operators at train and test time,
we introduce a variant of the CDP measurement scheme that depends on a scalar $\gamma \geq 1$ that controls the dynamic range of the mask magnitudes (and, as such, the conditioning of the sensing operator):
\begin{align}
    \mathcal{A}_{\gamma}(x) &= \begin{bmatrix}
        \mathcal{F}(D_{\gamma} (\mask_{1} \circ x)) \\
        \vdots \\
        \mathcal{F}(D_{\gamma} (\mask_{k} \circ x))
    \end{bmatrix}, \;
    D_{\gamma} = \diag(\set{e^{u_{j}}}_{j=1}^d),
    \label{eq:parameterized-family} \\
    \text{where} \;\; & u_{j} \iid \mathrm{Unif}([0, \log(\gamma)]).
    \notag
\end{align}
In particular, our experiments fix a particular $\gamma \equiv \gamma_{\text{train}}$ at train time and use datasets with $\gamma_{\text{test}} \neq \gamma_{\text{train}}$ at test time.
Note that the resulting measurement operator is no longer isometric: its condition number is bounded by
\begin{equation}
    1 \leq \kappa(\mathcal{A}_{\gamma}^*\mathcal{A}_{\gamma})
    \leq
    e^{2(\max_j u_j-\min_j u_j)}
    \leq \gamma^2.
    \label{eq:condnum-bound}
\end{equation}
Throughout, we write $A$ for the matrix induced by $\mathcal{A}_{\gamma}$; the particular value of $\gamma$ will be clear from the context:
\begin{equation}
    A = \begin{bmatrix}
        F_{d} D_{\gamma} \diag(\mask_1) \\
        \vdots \\
        F_{d} D_{\gamma} \diag(\mask_k)
    \end{bmatrix} \in \mathbb{C}^{kd \times d},
\end{equation}
where $F_{d}$ denotes the $d \times d$ orthogonal DFT matrix.

\subsection{Notation and basic constructions}
\label{sec:subsec:notation}
We write $\ip{\cdot, \cdot}$ for the Euclidean inner product, $\norm{x} := \sqrt{\ip{x, x}}$ for the Euclidean norm, and $\norm{x}_{M} := \sqrt{\ip{x, Mx}}$ for the weighted Euclidean norm induced by a positive-definite matrix $M$.
We write $\mathbb{S}^{d-1}$ for the unit sphere 
and $\bar{x}$ for the complex conjugate of $x$ (interpreted elementwise when $x$ is not scalar), and $A^*$ for the adjoint (Hermitian
transpose) of a complex matrix $A$.
We write $A^{\dagger}$ for the Moore-Penrose pseudoinverse of a matrix $A$ and $\opnorm{A} := \sup_{\norm{x} = 1} \norm{Ax}$ for its spectral norm.
The \emph{proximal operator} of a function $g$ is defined by
\begin{equation}
    \prox_{\lambda g}(\bar{x}) :=
    \argmin_{x} \set[\Big]{
        g(x) + \frac{1}{2\lambda} \norm{x - \bar{x}}^2
    }, \; \lambda > 0.
    \label{eq:prox}
\end{equation}

%% file: sections/method.tex
To recover $x_{\star}$, we aim to minimize the sum of a data-fidelity loss, such as the mean-squared error, and a regularizer $\mathcal{R}(x)$:
\begin{equation}
    \min_{x} \frac{1}{2} \norm{\abs{\mathcal{A}(x)} - y}^2 + \lambda \cdot \mathcal{R}(x),
    \label{eq:signal-recovery-optimization-problem}
\end{equation}
Following the regularization-by-denoising (RED) paradigm~\cite{REM17,RS19}, we use a deep denoiser $\mathcal{D}_{\theta}$ 
and set
\begin{equation}
    \mathcal{R}(x) = \frac{1}{2} \ip{x, x - \mathcal{D}_{\theta}(x)}.
    \label{eq:red-denoiser}
\end{equation}
Solving problems of the form $\min_{x} \set{f(x) + g(x)}$, such as the problem~\eqref{eq:signal-recovery-optimization-problem}, is commonly accomplished by \emph{proximal (sub)gradient} methods.
These methods iteratively minimize a local model of the objective function that comprises the sum of $g$ and the first-order Taylor expansion of $f$,
under the assumption that $\prox_{\lambda g}(x)$ is easily computable:
\begin{align*}
    x_{t+1} &= \argmin_{x} \set*{
        g(x) + f(x_t) + \ip{v_t, x - x_t} + \frac{1}{2 \lambda} \norm{x - x_t}^2
    } \\
    &= \prox_{\lambda g}(x_t - \lambda v_t), \;\;
    v_t \in \partial f(x_t),
\end{align*}
writing $\partial f(x_t)$ for the (Clarke) subdifferential~\cite{Clarke90} of $f$ at $x_t$.
When regularization involves a neural network, as in algorithm unrolling,
evaluating $\prox_{\lambda g}$ becomes less straightforward. In the regularization-by-denoising framework,
the authors of~\cite{MSVB18} provide a fixed-point characterization of the proximal mapping
that lends itself to a simple iterative approximation, 
but does not extend to generalized proximal mappings (\emph{i.e.}, where the standard Euclidean norm in~\eqref{eq:prox} is replaced with an arbitrary weighted Euclidean norm $\norm{\cdot}_{\Sigma}$) that are well-suited to preconditioned proximal gradient methods.
The forthcoming section addresses this setting.

\subsection{Alternating projections with a RED denoiser} \label{sec:subsec:alternating-projections-with-red}
In this section, we show how to embed a RED-type denoiser within the alternating projections method.
Our starting point is the observation (see, e.g.,~\cite[Section 4]{FS20}) that each step of the alternating projections method
is equivalent to preconditioned gradient descent with preconditioner $M := (A^* A)$:
\begin{align}
    \mathcal{T}_{\mathrm{AP}}(x_t) &=
    A^{\dag} \left(y \circ \phase(A x_t)\right)
    \label{eq:ap-as-preconditioned-gd}
    \\
    &= \argmin_{x \in \mathbb{R}^d} \set[\Big]{
        f(x_t) + \ip{v_t, x - x_t} + \frac{1}{2}
        \norm{x - x_t}_{M}^2
    }
    \notag \\
    &=
    x_{t} - M^{-1} v_t,
    \label{eq:ap-as-preconditioned-gd-ii}
\end{align}
where $\phase(z) := z / \abs{z}$ interpreted elementwise (with $\phase(0) = 0$), $f(x) = \frac{1}{2} \norm{\abs{\mathcal{A}(x)} - y}^2$ and $v_{t} \in \partial f(x_t)$.
Given the regularizer, we use a standard proximal update:
\begin{align}
    x_{t+1} &= \argmin_{x \in \mathbb{C}^d} \set[\Big]{
        Q(x; x_t) + \frac{1}{2} \norm{x - x_{t}}_{M}^2
    },
    \label{eq:ap-with-red} \\
    Q(x; x_t) &=
    f(x_t) + \ip{v_t, x - x_t} + \frac{\lambda}{2} \ip{x, x - \mathcal{D}_{\theta}(x)}.
    \notag
\end{align}
Under the standard local homogeneity and strong passivity conditions (cf.~\cite[Section 3.1]{REM17}),
the optimality condition of~\eqref{eq:ap-with-red} reduces to the nonlinear equation
\begin{equation}
    0 = v_{t} + M(x_{t+1} - x_{t}) + \lambda(x_{t+1} - \mathcal{D}_{\theta}(x_{t+1}))
    \label{eq:ap-red-optimality-condition}
\end{equation}
Rearranging the terms in~\eqref{eq:ap-red-optimality-condition} and using the identity~\eqref{eq:ap-as-preconditioned-gd-ii}
we obtain the following condition:
\begin{equation}
    x_{t+1} = (\lambda I + M)^{-1} \left[
    \lambda \mathcal{D}_{\theta}(x_{t+1}) + M \mathcal{T}_{\mathrm{AP}}(x_{t})
    \right].
    \label{eq:ap-red-fixed-point-equation}
\end{equation}
Treating~\eqref{eq:ap-red-fixed-point-equation} as a fixed-point equation, we propose approximating $x_{t+1}$ by a single step of the fixed-point iteration applied to $u \mapsto (\lambda I + M)^{-1}[\lambda \mathcal{D}_{\theta}(u) + M \mathcal{T}_{\mathrm{AP}}(x_{t})]$ initialized at $u_{0} := \mathcal{T}_{\mathrm{AP}}(x_{t})$.
Note that setting $M = I_{d}$ recovers the approximation used in~\cite[Section 3.1]{MSVB18} as a special case.
We collect the complete method in Algorithm~\ref{alg:ap-red}.

\begin{algorithm}[h]
    \caption{Alternating projections with RED denoising}
    \begin{algorithmic}
        \State \textbf{Inputs}: $x_{0} \in \mathbb{C}^d$, $\lambda > 0$, $T \in \mathbb{N}$
        \For{$t = 1, \dots, T$}
            \State $z_{t} = \mathcal{T}_{\mathrm{AP}}(x_{t-1})$
            \Comment{See~\eqref{eq:ap-as-preconditioned-gd}}
            \State $x_{t} = (\lambda I + A^* A)^{-1} \left[
                \lambda \mathcal{D}_{\theta}(z_t) + A^* A z_t
            \right]$
        \EndFor
        \State \Return $x_{T}$
    \end{algorithmic}
    \label{alg:ap-red}
\end{algorithm}

\subsection{Unrolled alternating projections} \label{subsubsec:unrolled-ap}
\begin{figure*}[ht]
    \centering
    \includegraphics[width=0.99\linewidth]{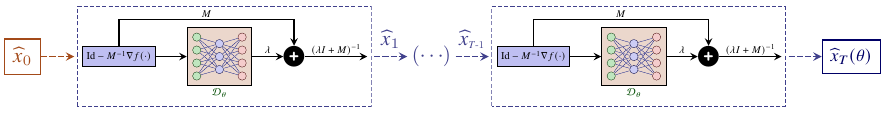}
    \caption{Overview of the proposed approach (\unrolledap): an initial estimate $\widehat{x}_0$ is iteratively refined through interleaved applications of the preconditioned gradient step $\mathcal{T}_{\mathrm{AP}}$ and the deep denoising operator $\mathcal{D}_{\theta}(\cdot;\sigma_t)$ (cf.~\eqref{eq:ap-red-fixed-point-equation}). The denoiser parameters are learned by minimizing the average squared reconstruction error $\norm{\widehat{x}_{T}(\theta) - x_{\star}}_{M}^2$.
    }
    \label{fig:unrolled-ap}
\end{figure*}

In this section, we describe the proposed method and architecture, termed \unrolledap; a graphical overview is provided in~\cref{fig:unrolled-ap}.
Given a training dataset of $n$
problem instances, $\mathcal{S}_{n} = \set{(x_{\star}^{(i)}, y^{(i)})_{i=1}^n}$,
we minimize a weighted least squares loss over denoiser parameters $\theta \in \mathbb{R}^p$:
\begin{equation}
    \mathcal{L}(\theta; \mathcal{S}_n) =
    \frac{1}{n} \sum_{i = 1}^{n}
    \norm{\widehat{x}_{T}^{(i)}(\theta) - x_{\star}^{(i)}}^2_{M},
    \;\; M := A^* A,
    \label{eq:unrolled-loss-function}
\end{equation}
Note that our training loss uses a weighted norm $\norm{\cdot}_{M}$ --- as opposed to the standard Euclidean norm --- since $M^{1/2}$ applies a whitening transform to the iterate space before computing the standard Euclidean loss.
Here $\widehat{x}_{T}^{(i)}(\theta)$ is the reconstruction obtained after $T$ unrolled iterations starting from the local initial point $x_0^{(i)}$ generated according to Section~\ref{sec:init}.

\medskip

\noindent \textbf{Denoiser architecture.}
In our implementation, we use the {\denoiser} architecture~\cite{ZLZ+21} with weight-tying
(i.e., using the same set of weights $\theta$ in each step of~\cref{alg:ap-red}),
following standard practice~\cite{MLE21}.
Since the {\denoiser} model requires a noise level parameter $\sigma$ as input, we learn an entire ``noise schedule'' $\set{\sigma_{t}}_{t=1}^{T}$ in an end-to-end fashion by setting
\begin{equation}
    x_{t} = (\lambda I + A^* A)^{-1} \left[
    \lambda \mathcal{D}_{\theta}(z_{t}; \sigma_{t}) + A^* A z_t
    \right].
    \label{eq:full-ap-update}
\end{equation}
In~\eqref{eq:unrolled-loss-function}, we abuse notation by using $\theta$ to refer to the concatenation of all learnable parameters;
i.e., both the denoiser weights and the set of learnable noise levels $\sigma_{t}$.

\subsection{Initialization}\label{sec:init}
Since the objective~\eqref{eq:signal-recovery-optimization-problem} is generally nonconvex, convergence to a global optimum is not a priori guaranteed. Our experiments employ two different initialization schemes:
\begingroup
\setlength{\leftmargini}{1em}
\begin{itemize}
    \item \textbf{synthetic initialization}: the initial signal is artificially chosen in a small ball around the ground truth to separate global search difficulties
    from local convergence behavior:
    \begin{equation}
        x_0 = x_{\star} + \delta \norm{x_{\star}} \cdot u, \;\; u \sim \mathrm{Unif}(\mathbb{S}^{d-1}).
        \label{eq:synthetic-init}
    \end{equation}
    \item \textbf{constant initialization}: a computationally inexpensive initialization method, suitable when $x_{\star}$ has a nontrivial DC component~\cite{MSVB18}, sets the initial estimate
    equal to
    \begin{equation}
        x_{0} = \mathbf{1}_{d} \cdot \frac{\norm{y}}{\norm{A}}.
        \label{eq:all-ones-init}
    \end{equation}
    The normalization constant ensures that $\norm{x_0} \approx \norm{x_{\star}}$.
\end{itemize}
\endgroup

%% file: sections/experiments.tex
To evaluate our method, we conduct synthetic experiments with measurements following~\eqref{eq:forward-model} corrupted by Poisson shot noise, varying the conditioning of the measurement operator via~\eqref{eq:parameterized-family}. We use the standard Gaussian/Poisson approximation
\[
	y^2 \sim \mathcal{N}(\abs{Ax}^2, \alpha^2 \diag(\abs{Ax}^2)),
\]
fixing the noise level to $\alpha = 27/255$.
Our evaluation metric is the relative error measured in the Euclidean norm induced by $M := A^*A$. Results are reported as mean $\pm$ standard deviation. We compare {\unrolledap} against two baselines: an unrolled variant of the proximal gradient method (using a learned scalar step-size and regularizer) termed {\trainedpg}, and a standard PnP scheme known as {\prdeep}~\cite{MSVB18}. We use $T = 10$ iterations for {\unrolledap} and {\trainedpg} and $T = 200$ for {\prdeep}, following the original recommendation.
Throughout, we fix $K = 4$ and $\gamma = 2.5$ for the training dataset.

\begin{figure*}
	\centering
	\begin{subfigure}[t]{0.3\textwidth}
		\centering
		\begin{tikzpicture}[baseline=(current axis.north)]
			\begin{axis}[
					width=0.85\linewidth,
					height=0.7\linewidth,
					scale only axis,
					ymode=log,
					xmode=log,
					ymax=9e-1,
					log basis x={2},
					grid=major,
					enlargelimits=0.02,
					xlabel style={yshift=-1pt},
					tick label style={font=\small},
					label style={font=\small},
					xlabel={Number of masks $K$},
					ylabel={$\norm{\widehat{x}_{T} - x_{\star}}_{M} / \norm{x_{\star}}_{M}$},
					xtick={1,2,4,8,16,32},
					legend style={
							at={(0.99,0.99)},
							anchor=north east,
							nodes={scale=0.65, transform shape}
						},
					legend entries={
							{\prdeep},
							{\trainedpg},
							{\unrolledap},
							{$\gamma=1.0$},
							{$\gamma=2.5$},
							{$\gamma=5.0$},
						},
				]
				\addlegendimage{prdeep, solid}
				\addlegendimage{proxgrad, solid}
				\addlegendimage{mredstyle, solid}
				\addlegendimage{thick, solid, color=black}
				\addlegendimage{thick, dashed, color=black}
				\addlegendimage{thick, dotted, color=black}

				\addplot[prdeep, solid] coordinates {
						(1,0.0693) +- (0.0,0.0145)
						(2,0.0227) +- (0.0,0.0040)
						(4,0.0074) +- (0.0,0.0014)
						(8,0.0045) +- (0.0,0.0009)
						(16,0.0031) +- (0.0,0.0006)
						(32,0.0021) +- (0.0,0.0004)
					};

				\addplot[proxgrad, solid] coordinates {
						(1,0.0049) +- (0.0,0.0008)
						(2,0.0038) +- (0.0,0.0006)
						(4,0.0026) +- (0.0,0.0004)
						(8,0.0020) +- (0.0,0.0003)
						(16,0.0018) +- (0.0,0.0003)
						(32,0.0018) +- (0.0,0.0003)
					};

				\addplot[mredstyle, solid] coordinates {
						(1,0.0049) +- (0.0,0.0008)
						(2,0.0036) +- (0.0,0.0005)
						(4,0.0025) +- (0.0,0.0003)
						(8,0.0018) +- (0.0,0.0002)
						(16,0.0013) +- (0.0,0.0002)
						(32,0.0010) +- (0.0,0.0001)
					};

				\addplot[prdeep, dashed,
					mark=square*,
					error bars/.cd,
					y dir=both,
					y explicit
				] coordinates {
						(1,0.3227) +- (0.0,0.0340)
						(2,0.1571) +- (0.0,0.0264)
						(4,0.0062) +- (0.0,0.0011)
						(8,0.0036) +- (0.0,0.0007)
						(16,0.0024) +- (0.0,0.0005)
						(32,0.0016) +- (0.0,0.0003)
					};

				\addplot[proxgrad, dashed] coordinates {
						(1,0.0231) +- (0.0,0.0035)
						(2,0.0209) +- (0.0,0.0037)
						(4,0.0129) +- (0.0,0.0017)
						(8,0.0078) +- (0.0,0.0009)
						(16,0.0047) +- (0.0,0.0005)
						(32,0.0033) +- (0.0,0.0003)
					};

				\addplot[mredstyle, dashed] coordinates {
						(1,0.0052) +- (0.0,0.0009)
						(2,0.0037) +- (0.0,0.0005)
						(4,0.0021) +- (0.0,0.0003)
						(8,0.0016) +- (0.0,0.0002)
						(16,0.0011) +- (0.0,0.0001)
						(32,0.0009) +- (0.0,0.0001)
					};

				\addplot[prdeep, dotted] coordinates {
						(1,0.3403) +- (0.0,0.0376)
						(2,0.2201) +- (0.0,0.0334)
						(4,0.0327) +- (0.0,0.0053)
						(8,0.0028) +- (0.0,0.0005)
						(16,0.0015) +- (0.0,0.0003)
						(32,0.0011) +- (0.0,0.0002)
					};

				\addplot[proxgrad, dotted] coordinates {
						(1,0.0268) +- (0.0,0.0045)
						(2,0.0282) +- (0.0,0.0056)
						(4,0.0188) +- (0.0,0.0033)
						(8,0.0130) +- (0.0,0.0017)
						(16,0.0094) +- (0.0,0.0013)
						(32,0.0042) +- (0.0,0.0005)
					};

				\addplot[mredstyle, dotted] coordinates {
						(1,0.0065) +- (0.0,0.0011)
						(2,0.0041) +- (0.0,0.0006)
						(4,0.0026) +- (0.0,0.0003)
						(8,0.0014) +- (0.0,0.0002)
						(16,0.0009) +- (0.0,0.0001)
						(32,0.0007) +- (0.0,0.0001)
					};

			\end{axis}
		\end{tikzpicture}
		\caption{
			Reconstruction error for varying number of masks $K$ (see~\cref{sec:subsec:reconstruction-error-vs-oversampling})
		}
		\label{fig:subfig:sweep_k_multi_g}
	\end{subfigure}\hfill
	\begin{subfigure}[t]{0.3\textwidth}
		\centering
		\begin{tikzpicture}[baseline=(current axis.north)]
			\begin{axis}[
					width=0.8\linewidth,
					height=0.7\linewidth,
					scale only axis,
					ymode=log,
					ymax=9e-1,
					grid=major,
					enlargelimits=0.02,
					xlabel style={yshift=-1pt},
					tick label style={font=\small},
					label style={font=\small},
					xlabel={Dynamic range factor $\gamma$},
					legend style={
							at={(0.99,0.99)},
							anchor=north east,
							nodes={scale=0.65, transform shape}
						},
				]

				\addplot[prdeep] coordinates {
						(1.0,0.0074) +- (0.0,0.0014)
						(1.5,0.0070) +- (0.0,0.0013)
						(2.0,0.0063) +- (0.0,0.0012)
						(2.5,0.0063) +- (0.0,0.0011)
						(3.0,0.0073) +- (0.0,0.0013)
						(3.5,0.0150) +- (0.0,0.0029)
					};
				\addlegendentry{\prdeep}

				\addplot[proxgrad] coordinates {
						(1.0,0.0594) +- (0.0,0.0079)
						(1.5,0.0769) +- (0.0,0.0087)
						(2.0,0.0393) +- (0.0,0.0045)
						(2.5,0.0503) +- (0.0,0.0057)
						(3.0,0.0806) +- (0.0,0.0080)
						(3.5,0.1445) +- (0.0,0.0118)
					};
				\addlegendentry{\trainedpg}

				\addplot[mredstyle] coordinates {
						(1.0,0.0031) +- (0.0,0.0003)
						(1.5,0.0026) +- (0.0,0.0003)
						(2.0,0.0023) +- (0.0,0.0003)
						(2.5,0.0023) +- (0.0,0.0003)
						(3.0,0.0022) +- (0.0,0.0003)
						(3.5,0.0023) +- (0.0,0.0003)
					};
				\addlegendentry{\unrolledap}
			\end{axis}
		\end{tikzpicture}
		\caption{
			Reconstruction error under measurement
			distribution shift (see~\cref{sec:subsec:distribution-shift-robustness}).
		}
		\label{fig:subfig:shift_robustness}
	\end{subfigure}\quad
	\begin{subfigure}[t]{0.3\textwidth}
		\centering
		\begin{tikzpicture}[baseline=(current axis.north)]
			\begin{axis}[
					width=0.8\linewidth,
					height=0.7\linewidth,
					scale only axis=true,
					ymode=log,
					xmode=log,
					log basis x={2},
					grid=major,
					enlargelimits=0.02,
					xlabel style={yshift=-1pt},
					tick label style={font=\small},
					label style={font=\small},
					xlabel={Unrolled steps $T$},
					legend style={
							at={(0.99,0.99)},
							anchor=north east,
							nodes={scale=0.65, transform shape}
						},
				]
				\addplot[mredstyle] coordinates {
						(2,0.0186) +- (0.0,0.0025)
						(4,0.0082) +- (0.0,0.0010)
						(8,0.0027) +- (0.0,0.0003)
						(16,0.0018) +- (0.0,0.0002)
						(32,0.0018) +- (0.0,0.0002)
						(64,0.0018) +- (0.0,0.0002)
					};
				\legend{{\unrolledap}}
			\end{axis}
		\end{tikzpicture}
		\caption{
			Reconstruction error as a function of the
			unrolling horizon $T$ (see~\cref{sec:subsec:reconstruction-error-vs-unrolling-steps}).
		}
		\label{fig:subfig:unroll_ablation}
	\end{subfigure}
	\caption{
		Reconstruction performance as a function of oversampling,
		dynamic range, and unrolling horizon.
	}
	\label{fig:ablation_studies}
\end{figure*}
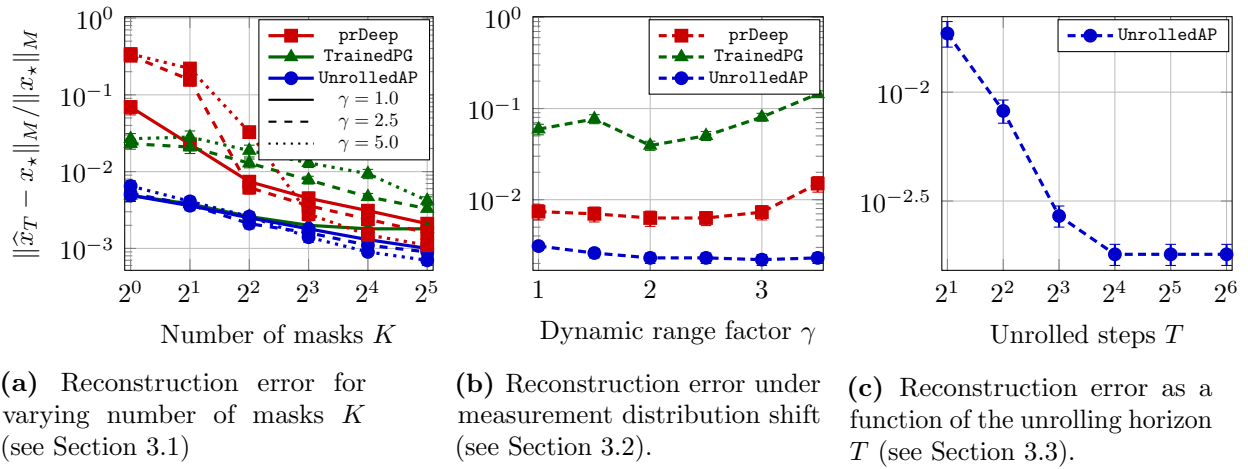

\vspace*{-0.5em}

\subsection{Effect of oversampling and conditioning}
\label{sec:subsec:reconstruction-error-vs-oversampling}
We first compare the proposed method and baselines under synthetic initialization,
varying the oversampling ratio $K \in \{1,2,4,8,16,32\}$
and the dynamic range factor $\gamma \in \{1.0, 2.5, 5.0\}$;
the results are illustrated in~\cref{fig:subfig:sweep_k_multi_g}.
Under perfect conditioning, all methods reduce the reconstruction error as $K$ increases, with {\unrolledap} achieving the lowest error across $K$.
The performance gap widens as the conditioning of the operator worsens.
Both baselines degrade for small and moderate $K$, while
while {\unrolledap} remains robust across oversampling ratios and condition numbers.

\subsection{Robustness to distribution shift} \label{sec:subsec:distribution-shift-robustness}

To evaluate the behavior of each method under distribution shift, we train {\unrolledap} and {\trainedpg} using datasets with $\gamma = 2.5$ and test across $\gamma \in \set{1.0,1.5,2.0,2.5,3.0,3.5}$.
As shown in \cref{fig:subfig:shift_robustness}, {\unrolledap} maintains a nearly constant error throughout, while both {\prdeep} and {\trainedpg} become more sensitive as the operator becomes ill-conditioned.
As expected, {\trainedpg} achieves the worst generalization among the three methods.

\subsection{Impact of unrolling steps $T$ on reconstruction error} \label{sec:subsec:reconstruction-error-vs-unrolling-steps}
To investigate the trade-off between computational efficiency and reconstruction accuracy, we conduct an ablation study on the number of unrolled steps $T = 2^k$ for $k \in \{1, \dots, 6\}$, using $\gamma = 2.5$ in both train and test datasets.
As shown in~\cref{fig:subfig:unroll_ablation} for the case of {\unrolledap}, the reconstruction error decreases rapidly as a function of $T$,
with diminishing returns for larger values $(T \geq 8)$.

\subsection{Performance across datasets}
\label{different_datasets}
We extend our evaluation to datasets with higher spatial complexity: \texttt{FashionMNIST} and (grayscale) \texttt{CIFAR10}. As illustrated in \cref{tab:dataset_comparison}, {\prdeep} performs well in the well-conditioned setting ($\gamma=1.0$), but degrades as the test condition number increases. Similarly, {\trainedpg} is sensitive to ill-conditioning with larger errors at higher condition numbers. In contrast, {\unrolledap} maintains low reconstruction errors across datasets and conditioning parameters.
This suggests that the preconditioning inherent in the alternating projections method is essential for stable reconstruction performance.

\begin{table}[ht]
	\centering
	\caption{Reconstruction error across datasets for models trained at $\gamma=2.5$%
		. All methods initialized according to~\eqref{eq:all-ones-init}.}
	\label{tab:dataset_comparison}
	\begin{tabular}{clccc}
		\toprule
		\textbf{Test $\gamma$} & \textbf{Method} & \textbf{MNIST}               & \textbf{FashionMNIST}        & \textbf{CIFAR-10}            \\
		\midrule
		\multirow{3}{*}{$\gamma=1.0$}
		                       & \prdeep         & $\mathbf{0.0083 \pm 0.0236}$ & $\mathbf{0.0066 \pm 0.0025}$ & $\mathbf{0.0050 \pm 0.0014}$ \\
		                       & \trainedpg      & $2.5296 \pm 1.9062$          & $0.3394 \pm 0.2533$          & $0.4114 \pm 0.2090$          \\
		                       & \unrolledap     & $0.0223 \pm 0.0078$          & $0.0210 \pm 0.0092$          & $0.0269 \pm 0.0280$          \\
		\midrule
		\multirow{3}{*}{$\gamma=2.5$}
		                       & \prdeep         & $1.0051 \pm 0.0867$          & $0.6865 \pm 0.1823$          & $0.3088 \pm 0.1131$          \\
		                       & \trainedpg      & $1.0082 \pm 0.1737$          & $1.1207 \pm 0.2734$          & $1.2190 \pm 0.4160$          \\
		                       & \unrolledap     & $\mathbf{0.0181 \pm 0.0122}$ & $\mathbf{0.0171 \pm 0.0115}$ & $\mathbf{0.0129 \pm 0.0061}$ \\
		\midrule
		\multirow{3}{*}{$\gamma=5.0$}
		                       & \prdeep         & $1.2363 \pm 0.0535$          & $1.1515 \pm 0.0675$          & $1.0446 \pm 0.0355$          \\
		                       & \trainedpg      & $1.5375 \pm 0.4126$          & $2.9983 \pm 0.2756$          & $3.7331 \pm 0.1386$          \\
		                       & \unrolledap     & $\mathbf{0.0508 \pm 0.0475}$ & $\mathbf{0.0324 \pm 0.0358}$ & $\mathbf{0.0296 \pm 0.0158}$ \\
		\midrule
		\multirow{3}{*}{$\gamma=10.0$}
		                       & \prdeep         & $1.4396 \pm 0.0249$          & $1.3621 \pm 0.0381$          & $1.3669 \pm 0.0203$          \\
		                       & \trainedpg      & $3.5381 \pm 0.2832$          & $5.3103 \pm 0.1897$          & $5.8642 \pm 0.0955$          \\
		                       & \unrolledap     & $\mathbf{0.0841 \pm 0.0734}$ & $\mathbf{0.0568 \pm 0.0498}$ & $\mathbf{0.0558 \pm 0.0525}$ \\
		\bottomrule
	\end{tabular}
\end{table}

%% file: sections/conclusion.tex
\section{Conclusion and future work}
We proposed an end-to-end unrolled alternating projections method that adapts to the geometry of the measurement operator with a learned RED-based denoiser for the phase retrieval problem.
Our work naturally leads to several possible future directions: for example, this paper empirically establishes robustness of the proposed scheme to changes in the dynamic range of the measurement operator, which does not cover the variety of different types of distribution shift considered in inverse problems.
Similarly, another direction would be to establish theoretical convergence guarantees under controlled shifts of the forward model.
Finally, we highlight that a particularly appealing aspect of our method is that it eschews the use of explicit model adaptation schemes such as transfer learning or retraining; consequently, the design of ``application-agnostic'' algorithmic schemes with similar robustness properties is a promising future direction.

%% file: main.bib
@article{MLE21,
  title={Algorithm unrolling: Interpretable, efficient deep learning for signal and image processing},
  author={Monga, Vishal and Li, Yuelong and Eldar, Yonina C},
  journal={IEEE Signal Processing Magazine},
  volume={38},
  number={2},
  pages={18--44},
  year={2021},
  publisher={IEEE}
}

@article{REM17,
  title={The little engine that could: Regularization by denoising {(RED)}},
  author={Romano, Yaniv and Elad, Michael and Milanfar, Peyman},
  journal={SIAM Journal on Imaging Sciences},
  volume={10},
  number={4},
  pages={1804--1844},
  year={2017},
  publisher={SIAM}
}

@article{FS20,
  title={The numerics of phase retrieval},
  author={Fannjiang, Albert and Strohmer, Thomas},
  journal={Acta Numerica},
  volume={29},
  pages={125--228},
  year={2020},
  publisher={Cambridge University Press}
}

@InProceedings{MSVB18,
  title = 	 {pr{D}eep: Robust Phase Retrieval with a Flexible Deep Network},
  author =       {Metzler, Christopher and Schniter, Phillip and Veeraraghavan, Ashok and Baraniuk, Richard},
  booktitle = 	 {Proceedings of the 35th International Conference on Machine Learning},
  pages = 	 {3501--3510},
  year = 	 {2018},
  editor = 	 {Dy, Jennifer and Krause, Andreas},
  volume = 	 {80},
  series = 	 {Proceedings of Machine Learning Research},
  month = 	 {10--15 Jul},
  publisher =    {PMLR},
  url = 	 {https://proceedings.mlr.press/v80/metzler18a.html},
}

@article{ZLZ+21,
  title={Plug-and-play image restoration with deep denoiser prior},
  author={Zhang, Kai and Li, Yawei and Zuo, Wangmeng and Zhang, Lei and Van Gool, Luc and Timofte, Radu},
  journal={IEEE Transactions on Pattern Analysis and Machine Intelligence},
  volume={44},
  number={10},
  pages={6360--6376},
  year={2021},
  publisher={IEEE}
}

@InProceedings{UVL18,
author = {Ulyanov, Dmitry and Vedaldi, Andrea and Lempitsky, Victor},
title = {Deep Image Prior},
booktitle = {Proceedings of the IEEE Conference on Computer Vision and Pattern Recognition (CVPR)},
month = {June},
year = {2018}
}

@INPROCEEDINGS{VBW13,
  author={Venkatakrishnan, Singanallur V. and Bouman, Charles A. and Wohlberg, Brendt},
  booktitle={2013 IEEE Global Conference on Signal and Information Processing}, 
  title={Plug-and-Play priors for model based reconstruction}, 
  year={2013},
  volume={},
  number={},
  pages={945-948},
  doi={10.1109/GlobalSIP.2013.6737048}}

@article{KBBW23,
  title={Plug-and-play methods for integrating physical and learned models in computational imaging: Theory, algorithms, and applications},
  author={Kamilov, Ulugbek S and Bouman, Charles A and Buzzard, Gregery T and Wohlberg, Brendt},
  journal={IEEE Signal Processing Magazine},
  volume={40},
  number={1},
  pages={85--97},
  year={2023},
  publisher={IEEE}
}

@article{AMMG19,
  title={Learning step sizes for unfolded sparse coding},
  author={Ablin, Pierre and Moreau, Thomas and Massias, Mathurin and Gramfort, Alexandre},
  journal={Advances in Neural Information Processing Systems},
  volume={32},
  year={2019}
}

@article{GS72,
  title={A practical algorithm for the determination of plane from image and diffraction pictures},
  author={Gerchberg, Ralph W},
  journal={Optik},
  volume={35},
  number={2},
  pages={237--246},
  year={1972}
}

@article{Fienup82,
  title={Phase retrieval algorithms: a comparison},
  author={Fienup, James R},
  journal={Applied optics},
  volume={21},
  number={15},
  pages={2758--2769},
  year={1982},
  publisher={Optical Society of America}
}

@article{gilton2021model,
  title={Model adaptation for inverse problems in imaging},
  author={Gilton, Davis and Ongie, Gregory and Willett, Rebecca},
  journal={IEEE Transactions on Computational Imaging},
  volume={7},
  pages={661--674},
  year={2021},
  publisher={IEEE}
}

@book{Clarke90,
  author    = {Clarke, Frank H.},
  publisher = {Society for Industrial and Applied Mathematics},
  title     = {{Optimization and Nonsmooth Analysis}},
  year      = {1990},
  doi       = {10.1137/1.9781611971309},
}

@article{KMW17,
  author={Kamilov, Ulugbek S. and Mansour, Hassan and Wohlberg, Brendt},
  journal={IEEE Signal Processing Letters}, 
  title={A Plug-and-Play Priors Approach for Solving Nonlinear Imaging Inverse Problems}, 
  year={2017},
  volume={24},
  number={12},
  pages={1872-1876},
  doi={10.1109/LSP.2017.2763583}}

@ARTICLE{RS19,
  author={Reehorst, Edward T. and Schniter, Philip},
  journal={IEEE Transactions on Computational Imaging}, 
  title={Regularization by Denoising: Clarifications and New Interpretations}, 
  year={2019},
  volume={5},
  number={1},
  pages={52-67},
  doi={10.1109/TCI.2018.2880326}}

@inproceedings{MME19,
author = {Mataev, Gary and Milanfar, Peyman and Elad, Michael},
title = {DeepRED: Deep Image Prior Powered by RED},
booktitle = {Proceedings of the IEEE/CVF International Conference on Computer Vision (ICCV) Workshops},
month = {Oct},
year = {2019}
}

@ARTICLE{OJM+20,
  author={Ongie, Gregory and Jalal, Ajil and Metzler, Christopher A. and Baraniuk, Richard G. and Dimakis, Alexandros G. and Willett, Rebecca},
  journal={IEEE Journal on Selected Areas in Information Theory}, 
  title={Deep Learning Techniques for Inverse Problems in Imaging}, 
  year={2020},
  volume={1},
  number={1},
  pages={39-56},
  doi={10.1109/JSAIT.2020.2991563}}
